\documentclass[journal]{IEEEtran}

\usepackage{times}
\usepackage{soul}
\usepackage{url}
\usepackage[colorlinks]{hyperref}
\usepackage[utf8]{inputenc}
\usepackage[small]{caption}
\usepackage{graphicx}
\usepackage{amsmath, amssymb}
\usepackage{amsthm}
\usepackage{booktabs}
\usepackage{algorithm}
\usepackage{algorithmic}
\usepackage[switch]{lineno}
\usepackage{tabularx}
\usepackage{booktabs}
\usepackage{textcomp}
\usepackage{xcolor}
\usepackage{multirow}
\usepackage{pifont} 
\usepackage{amssymb} 
\usepackage{pifont}
\usepackage{ragged2e}
\usepackage{hyperref}
\usepackage{footnote}

\usepackage[nospace]{cite}
\usepackage{tabularx} 
\newcommand{\eg}{{\em e.g.,~}}

\begin{document}

\title{Exploring Federated Unlearning: Review, Comparison, and Insights }

\author{Yang~Zhao,~\IEEEmembership{Member,~IEEE,}
        Jiaxi~Yang,~\IEEEmembership{Student Member,~IEEE,}
        Yiling~Tao,~\IEEEmembership{Student Member,~IEEE,}
        Lixu~Wang,~\IEEEmembership{Student Member,~IEEE,}
        Xiaoxiao~Li,~\IEEEmembership{Member,~IEEE,}
        Dusit~Niyato,~\IEEEmembership{Fellow,~IEEE,}
        and~H.~Vincent~Poor,~\IEEEmembership{Life Fellow,~IEEE}



\thanks{Y. Zhao and J. Yang contributed equally to this work.}
\thanks{Y. Zhao and D. Niyato are with Nanyang Technological University, Singapore (e-mail: s180049@e.ntu.edu.sg, dniyato@ntu.edu.sg).}
\thanks{J. Yang. (e-mail: abbottyang@std.uestc.edu.cn).}
\thanks{Y. Tao. (e-mail: 202164010387@mail.scut.edu.cn).}
\thanks{L. Wang is with Northwestern University, USA (e-mail: lixuwang2025@u.northwestern.edu).}
\thanks{X. Li is with the University of British Columbia, Canada (e-mail: xiaoxiao.li@ece.ubc.ca).}
\thanks{H. V. Poor is with the Department of Electrical and Computer Engineering, Princeton University, Princeton, NJ08544, USA (e-mail: poor@princeton.edu).}
}

\maketitle

\begin{abstract}
The increasing demand for privacy-preserving machine learning has spurred interest in federated unlearning, which enables the selective removal of data from models trained in federated systems. However, developing federated unlearning methods presents challenges, particularly in balancing three often conflicting objectives: privacy, accuracy, and efficiency. This paper provides a comprehensive analysis of existing federated unlearning approaches, examining their algorithmic efficiency, impact on model accuracy, and effectiveness in preserving privacy. We discuss key trade-offs among these dimensions and highlight their implications for practical applications across various domains. Additionally, we propose the \textsc{OpenFederatedUnlearning} framework, a unified benchmark for evaluating federated unlearning methods, incorporating classic baselines and diverse performance metrics. Our findings aim to guide practitioners in navigating the complex interplay of these objectives, offering insights to achieve effective and efficient federated unlearning. Finally, we outline directions for future research to further advance the state of federated unlearning techniques.
\end{abstract}


\IEEEpeerreviewmaketitle

\maketitle

\section{Introduction}

Federated learning is a distributed paradigm in which clients independently train a model on their local datasets and subsequently transmit updates, rather than raw data, to a central server. The server aggregates these updates into a global model, which is then shared back to the participating clients for further iterations. Despite preserving data locally, this process can unintentionally disclose sensitive information through the gradient updates exchanged between clients and the server, thus enabling membership inference attacks. Such vulnerabilities compromise user privacy, and conflict with regulatory requirements, underscoring the need for robust safeguards in federated learning environments. Moreover, with the emergence of new data regulations such as the European Union's General Data Protection Regulation (GDPR), clients are granted the ``right to be forgotten", enabling them to request the removal of their private data from models even after the data have been utilized for training a global model. A straightforward approach to accomplishing this is retraining the model from scratch without the removed data. However, this approach imposes a considerable computational burden. Additionally, when implementing unlearning in a decentralized manner, additional challenges arise. First, how to minimize the effect of unlearning on the accuracy of the federated learning model? Second, how can we guarantee that the client’s data is effectively erased as requested? Federated unlearning is crucial for removing obsolete data to keep models relevant and accurate in dynamic environments, such as personalized recommendations or autonomous systems.

\begin{figure}[!ht]
    \centering
    \includegraphics[scale=0.4]{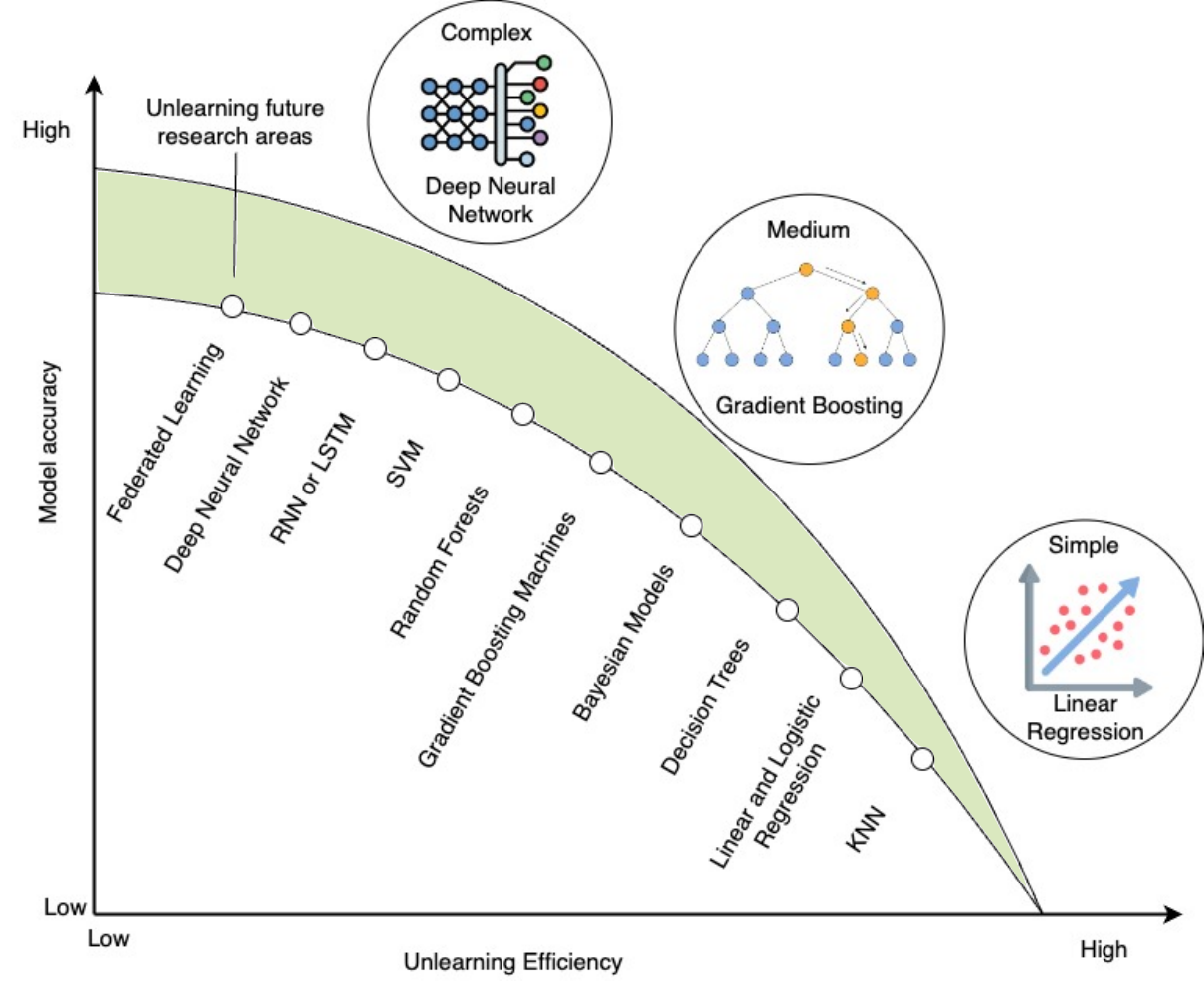}
    \caption{Trade-offs between unlearning efficiency and model accuracy. As a machine learning model's complexity increases, its accuracy increases while its unlearning efficiency decreases.}
    \label{fig:tradeoff}
\end{figure}

\begin{figure*}[!h]
    \centering
    \includegraphics[scale=0.32]{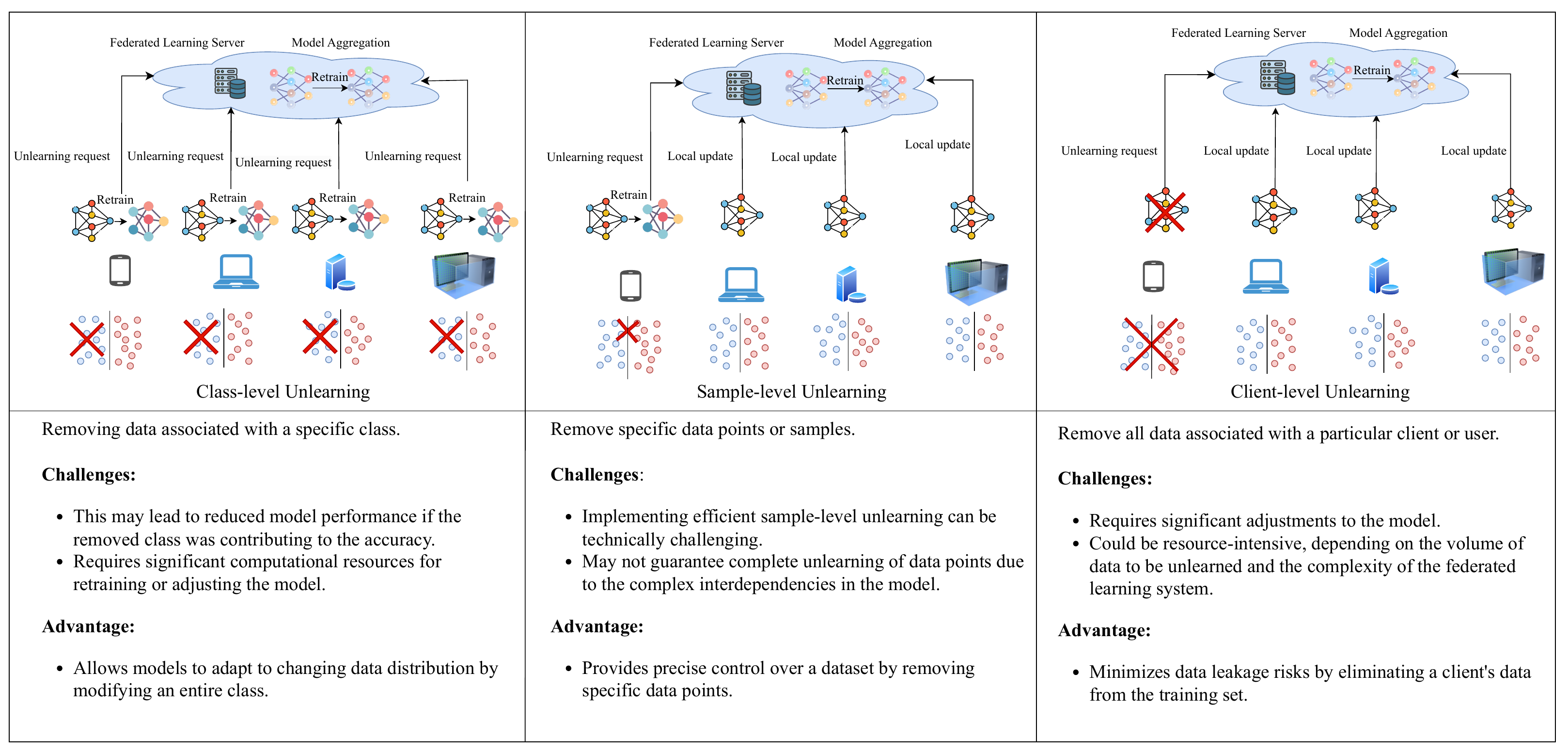}
    \caption{Federated unlearning workflows in class-level unlearning, sample-level unlearning, and client-level unlearning scenarios. Each flow includes removing data from clients' datasets, locally updating models, aggregating updated models to revise the global model, and validating model accuracy before redeployment.}
    \label{fig:fu-system}
\end{figure*}

Furthermore, we need to appreciate the trade-offs of unlearning. For example, the trade-off between the efficiency and accuracy of a model is shown in Figure~\ref{fig:tradeoff}. Usually, as the model’s complexity increases, its accuracy improves because more complex models, for example, deep neural networks, can better represent latent structures in complicated datasets, provided that they are not overfitted to a specific task. In the case of simpler models such as K-Nearest Neighbor, unlearning is simpler since simple models have fewer parameters and the relation between data and outputs is easy to perceive. However, in complex models, removing particular data points affects many parameters and layers making the unlearning methods computationally intensive. Besides, data are distributed among the clients' devices in federated learning, which brings out additional difficulties in effectively employing unlearning without deteriorating the model’s accuracy.


Federated unlearning addresses the need to selectively remove data from federated learning systems in order to comply with privacy regulations. Unlike conventional techniques (e.g., device anonymity, secure aggregation, differential privacy, and SMPC) that safeguard data during training but lack retrospective removal capabilities, federated unlearning enables fine-grained erasure while preserving model integrity. As illustrated in Figure~\ref{fig:fu-system}, federated unlearning requests typically target three levels: class-level (removing all data within a class), client-level (removing all data originating from a specific client), and sample-level (removing the influence of particular data points). Notably, while the federated unlearning framework supports the removal of client data, it specifically focuses on mitigating the residual impact of such data on the learned model rather than physically deleting raw data from client devices. Numerous federated unlearning methods have been proposed, and we provide insights from the following perspectives :

\noindent \textbf{Major Applications of Federated Unlearning:}
What are the primary federated unlearning scenarios and domains, and what specific problems federated unlearning can solve within these contexts?

\noindent \textbf{High-Performance Federated Unlearning Methods:}
What are the critical considerations and design principles that should be taken into account while developing federated unlearning methods?

\noindent \textbf{Threats in Federated Unlearning:}
What potential security threats and risks may result from adopting federated unlearning?

\noindent \textbf{Trade-Offs in Federated Unlearning:}
What are the trade-offs that need to be considered when applying federated unlearning, particularly regarding the trade-off between model accuracy, data privacy, and efficiency?

\noindent \textbf{Metrics for Evaluating Federated Unlearning:}
Which metrics are most effective in assessing the performance of federated unlearning methods, especially in terms of accuracy and efficiency of unlearning? How to verify that data removal requirements have been successfully met?

In this study, we compare existing methods and focus on analyzing trade-offs between the accuracy and efficiency of unlearning processes. For instance, the authors in~\cite{wu2022federated} presented a comprehensive overview of federated unlearning methods development. On the contrary, the authors in~\cite{wang2023federated} identified privacy risks related to federated unlearning. Additionally, the authors in~\cite{guo2023fast} suggested performing the server-level federated unlearning by eliminating harmful devices. Other authors~\cite{wang2022federated, su2023asynchronous, yuan2023federated, zhu2023heterogeneous, wang2023bfu, xia2023fedme, zhang2023fedrecovery, xiong2023exact, che2023fast, zhao2023federated, liu2021federaser} have contributed to the development of diverse federated unlearning methods aimed at mitigating negative impacts on federated learning systems. Furthermore, the authors in~\cite{wang2023edge} have adopted federated unlearning methods for Vehicle-to-Everything (V2X) communications applications.

\textbf{Contributions.} To the best of our knowledge, this is the first comprehensive article to present a detailed comparison and evaluation of federated unlearning methods. We focus on exploring the trade-offs between accuracy, efficiency, and privacy in federated unlearning. Additionally, we provide a comparative analysis of existing methods, highlighting their benefits and limitations, and propose evaluation metrics to assess their effectiveness in ensuring data removal compliance while maintaining model integrity and computational efficiency. Furthermore, we introduce \textsc{OpenFederatedUnlearning}, a unified benchmark framework that incorporates classic baselines and diverse performance metrics for evaluating federated unlearning approaches. Based on experimental results, we offer valuable guidelines for implementation and propose directions for future research to advance the development of federated unlearning methods.

\section{Federated Unlearning Approaches, Applications, and Challenges}
\label{sec:trustworthy}

Prevalent data regulations, such as the GDPR, underscore the need for machine unlearning methods~\cite{liu2021federaser}. These regulations aim to uphold individual privacy rights, particularly the ``right to be forgotten" by enabling a selective erasure of user data from models. With federated learning, a privacy-enhancing machine learning algorithm, more advanced unlearning methods, i.e., federated unlearning, are required to maintain model accuracy after data removal, especially on network edge devices with limited computational resources~\cite{liu2021federaser, wang2023bfu, xiong2023exact}. Additionally, unlearning provides safeguards against malicious threats and manages challenges posed by independent and identically distributed (non-IID) and poisoned data~\cite{liu2021federaser}. Furthermore, federated unlearning enhances federated learning systems by facilitating privacy-preserving knowledge exchange and ensuring that models can adapt by incorporating fresh information and discarding obsolete data~\cite{guo2023fast}.

In this section, we introduce a variety of federated unlearning approaches, including training-based, tuning-based, and model-based methods, and their practical applications across critical fields such as healthcare and finance, where strict adherence to privacy is essential. Table~\ref{table-all} highlights the benefits, limitations, and applications of these approaches, illustrating how each method aligns with specific requirements and challenges in federated learning environments.

\subsection{Approaches}
Federated unlearning methods are classified into three approaches: training-based, tuning-based, and model-based. Each approach can be applied at different levels of granularity: class-level, client-level, and sample-level, depending on the scope of the data to be forgotten.

\textbf{Training-Based Approaches} focus on retraining the model with the remaining dataset after removing specific data points. This ensures complete removal of the influence of the forgotten data but comes with significant computational costs. For instance, the FedME$^2$ framework~\cite{xia2023fedme} and Fast Federated Model Unlearning (FFMU) framework~\cite{che2023fast} involve re-evaluating the model's memory or refining it with adjusted data. These methods aim to preserve privacy while minimizing losses in model accuracy at the class, client, or sample level. The key advantage of training-based approaches is their ability to guarantee full compliance with data removal requests and eliminate the residual influence of forgotten data. However, their significant computational cost makes them impractical for large-scale systems, especially in dynamic federated environments.

\textbf{Tuning-Based Approaches} adjust the parameters of a pre-trained model to selectively forget specific data contributions. These methods are typically more efficient than full retraining and are well-suited for client-level and sample-level unlearning. For example, Reverse SGA~\cite{wu2022federated} and FedEraser~\cite{liu2021federaser} illustrate how parameter adjustments can effectively remove data influence while maintaining overall model accuracy.  The main benefit of tuning-based approaches is their computational efficiency and their ability to maintain overall model performance. However, these methods require careful implementation to ensure complete data removal and may leave residual traces of forgotten data if not rigorously applied, posing risks in privacy-sensitive applications. 

\textbf{Model-Based Approaches} modify the model structure directly to address data forgetting. These methods are particularly effective for class-level unlearning, where structural changes such as pruning or removing specific components ensure that an entire class of data is eliminated. For instance, TF-IDF quantization and pruning methods within Convolutional Neural Networks (CNNs)~\cite{wang2022federated} directly adjust the model's structure to eliminate the influence of specific data classes.  The primary advantage of model-based approaches is their efficiency in specific scenarios such as class-level unlearning, where they can ensure the precise removal of targeted data. However, these methods may impact the model's adaptability and generalization, potentially limiting their applicability in diverse or continuously evolving federated learning environments.

\subsection{Applications}

Federated unlearning's ability to manage data and ensure regulatory compliance makes it applicable across various domains, each with specific requirements:

\begin{itemize}

\item\textbf{Telecommunications and Smart City.} Federated unlearning is crucial in telecommunications and smart city applications, addressing scalability, real-time processing, and the diversity of data sources. FedFilter~\cite{wang2023edge} framework enhances secure communications by enabling the selective erasure of outdated or incorrect data, thereby optimizing system performance and trustworthiness. In particular, it uses edge caching to expedite federated unlearning by offloading model updates closer to the data, and applies user‐clustering techniques (e.g., sparsity‐based similarity) to deliver targeted personalization. Its hierarchical aggregation strategy shares only ``base” model layers globally, keeping specialized ``personalized” layers local to each cluster, which lowers communication overhead and mitigates the risk of data leakage. The unlearning component then removes malicious or obsolete inputs directly at or near the edge, curbing any negative impact on system accuracy and security. By localizing these processes, FedFilter reduces privacy risks and latency for real‐time IoV operations in smart cities, ensuring robust performance even in large‐scale, heterogeneous environments~\cite{wang2023edge}.

\item \textbf{Healthcare and Finance.} Data confidentiality and regulatory compliance are crucial in healthcare and finance. For example, the FedEraser framework supports the secure removal of sensitive client data from federated learning models, mitigating breach risks and ensuring privacy regulation compliance~\cite{liu2021federaser}. Effective data governance is vital for maintaining trust and preventing attacks.

\item\textbf{The Internet of Things (IoT).} Federated learning-based IoT systems involve numerous interconnected devices that generate and process data, requiring low-power solutions for removing malicious and obsolete data. In particular, FRU optimizes storage efficiency on resource-constrained devices by storing historical updates locally and employing negative sampling and importance-based update selection~\cite{yuan2023federated}. This enables on-device unlearning, efficient storage utilization, enhanced privacy by erasing user data, and rapid model reconstruction for continuous services in IoT environments.

\item \textbf{Knowledge Graph Embedding.} Federated unlearning has potential applications in knowledge graph embedding, such as FedLU introduces an unlearning approach inspired by cognitive neuroscience, combining retroactive interference and passive decay to selectively erase specific knowledge from local embeddings and propagate this forgetting to the global model~\cite{zhu2023heterogeneous}.
\end{itemize}

\subsection{Challenges}
\label{sec:efficiency}

Advanced federated unlearning aims to efficiently remove data from models without requiring retraining, while addressing challenges related to efficiency, privacy, accuracy, scalability, adaptability, and robustness~\cite{wu2022federated, zhang2023fedrecovery, wang2023federated, wang2023bfu, xia2023fedme, su2023asynchronous, zhu2023heterogeneous}. These approaches leverage privacy-preserving models, manage trade-offs between data removal and model accuracy, and tackle security risks such as data poisoning and membership inference attacks~\cite{wang2023federated, xiong2023exact}. Techniques such as differential privacy~\cite{zhang2023fedrecovery}, historical parameter usage~\cite{liu2021federaser}, and cluster-based aggregation~\cite{su2023asynchronous} are used to optimize unlearning efficiency, enhance scalability, and maintain model performance while ensuring privacy protection. The key goal is to achieve effective data removal with minimal impact on model accuracy, enabling federated learning systems to remain adaptive and secure in diverse, large-scale environments.
\begin{itemize}
\item \textbf{Maintain Model Accuracy.} In federated unlearning, achieving both data removal and model accuracy is key. FedME$^2$ uses a Memory Evaluation (MErase) module to determine which data to unlearn, minimizing accuracy loss~\cite{xia2023fedme}. By fine-tuning model weights, FedME$^2$ ensures gradual mitigation of unlearned data impact, with accuracy impacts kept under $4\%$ at a $75\%$ data forgetting rate. Similarly, the Bayesian Federated Unlearning (BFU) algorithm leverages Bayesian methods to update models based on remaining and erased data, eliminating the accuracy-unlearning trade-off~\cite{wang2023bfu}. BFU-SS achieves $97.68\%$ accuracy on MNIST while unlearning $10\%$ of data.

\item \textbf{Efficient Unlearning.} In addition to model accuracy, efficiency is a key metric for evaluating federated unlearning algorithms. Effective approaches should swiftly update global models while minimizing energy use and avoiding the overhead of retraining. FedRecovery employs differential privacy with tailored Gaussian noise to align unlearned models with retrained models, reducing computational costs~\cite{zhang2023fedrecovery}. It achieves processing times of approximately $1$ second compared to $679.1$ seconds for retraining. Similarly, FedEraser leverages historical updates to restore models efficiently~\cite{liu2021federaser}. This approach yields a $3.4$× reduction in unlearning time for the Adult dataset, and a $4.8$× and $4.1$× reduction for the MNIST and Purchase datasets, respectively.

\item \textbf{Adaptation to Real-Time Data Dynamics.} Efficient federated unlearning algorithms also emphasize the capability of adaptation, since models are changing constantly in response to real-time data dynamics, such as evolving data distributions, as well as removing outdated and wrong data. This is demonstrated in the TF-IDF algorithm used for channel discrimination in CNNs~\cite{wang2022federated}. It effectively adjusts the model to the updated dataset without extensive retraining, focusing on channels relevant to the data categories to be unlearned. A subsequent fine-tuning step mitigates any accuracy loss, ensuring the model remains accurate and aligned with the new data distribution after unlearning. In terms of efficiency, TF-IDF significantly speeds up the unlearning process by $8.9$× for the ResNet model and $7.9$× for the Visual Geometry Group (VGG) model on the CIFAR-10 dataset, all without any loss of accuracy, in comparison to training the models from scratch. At the same time, FedLU deals with the discrepancy between local optimization and global convergence by supporting local models to independently forget and aggregate global information. This enables local models to adjust by forgetting outdated knowledge and controlling the trade-off between learning new data and unlearning obsolete knowledge~\cite{zhu2023heterogeneous}.

\item \textbf{Mitigating Security Risks.} Federated unlearning should ensure efficient data removal while safeguarding privacy. Variations in model versions before and after unlearning can make the system vulnerable to data poisoning threats, such as Membership Inference Attacks (MIAs), Model Inversion Attacks, and Gradient Leakage Attacks~\cite{wang2023federated}. To mitigate MIAs, Exact-Fun employs gradient descent with noise addition during unlearning to obscure the influence of individual data samples~\cite{xiong2023exact}. This added noise reduces traceability, enhancing data protection.

\item \textbf{Scalability in Large-Scale Systems.} In large-scale federated learning, federated unlearning must avoid overloading the network or central server, ensuring timely updates aligned with privacy requirements. The FFMU algorithm, designed for resource-limited edge devices, employs tools such as the Nemytskii operator for efficient local unlearning~\cite{che2023fast}. FFMU supports concurrent training and unlearning, optimizing local models to reduce computational demands and enhance scalability. The KNOT framework utilizes a clustered aggregation approach, focusing on specific clusters for unlearning to minimize retraining costs and reduce computational overhead~\cite{su2023asynchronous}. Together, these methods ensure efficient, scalable unlearning without burdening the system.
\end{itemize}

In summary, tailored noise is a very efficient method for dealing with security threats of federated unlearning algorithms. Moreover, storing historical parameters can effectively speed up the federated unlearning without retraining from scratch. Furthermore, federated unlearning is also applied to delete obsolete data and thereby ensure the actuality of federated learning systems.

\begin{table*}[]
\centering
\begin{tabular}{|>{\centering\arraybackslash}p{0.1\textwidth}|>{\centering\arraybackslash}p{0.1\textwidth}|>{\centering\arraybackslash}p{0.2\textwidth}|>{\centering\arraybackslash}p{0.15\textwidth}|>{\centering\arraybackslash}p{0.15\textwidth}|>{\centering\arraybackslash}p{0.15\textwidth}|}
\hline
\textbf{Approach} & \textbf{Granularity} & \textbf{Algorithm} & \textbf{Application} & \textbf{Benefits} & \textbf{Limitations} \\ \hline

Training-Based & Client-Level & MoDe aligns parameters for model knowledge erasure, with memory guidance for maintaining model accuracy~\cite{zhao2023federated}. & Efficient in federated learning for client revocation and category removal. & Improves data privacy by unlearning specific data points. & Accuracy degradation due to parameters update. \\ \hline

Tuning-Based & Client-Level & Asynchronous federated unlearning using clustered aggregation and optimized client-cluster assignment~\cite{su2023asynchronous}. & Allows for the selective removal of data samples with minimal retraining costs. & Reduces retraining costs by focusing on affected clusters. & Scalability to large federated learning systems may be challenging. \\ \hline

Tuning-Based & Client-Level & FedRecovery masks the gap between unlearned and retrained models with a Gaussian mechanism~\cite{zhang2023fedrecovery}. & Improves privacy and efficiency of federated learning settings by eliminating sensitive attributes. & Efficiently reproduces a model indistinguishable from the retrained one. & Gaussian noise may impact model accuracy. \\ \hline

Tuning-Based & Client-Level & FedEraser leverages historical parameter updates for specific client data removal~\cite{liu2021federaser}. & Useful in sensitive domains including healthcare and finance for enhancing data privacy. & FedEraser offers efficient data removal, time savings in model reconstruction, and non-intrusive operation within existing federated learning systems. & Possible supplementary storage needs, implementation difficulty, and scalability issues over a large number of clients and complex models. \\ \hline

Training-Based & Client-Level & FAST eliminates contributions of malicious clients, judging unlearning efficiency, and remedying unlearning model accuracy~\cite{guo2023fast}. & Enhances security and robustness in sectors such as healthcare and finance. & Improves model integrity by removing malicious data. & Removes malicious contributions can require extra resources. \\ \hline

Training-Based & Client-Level & FRU revises updates for efficient unlearning in federated recommendation systems~\cite{yuan2023federated}. & Enhances data privacy and system robustness by erasing compromised data. & Optimizes storage on client devices with innovative strategies. & Unlearning may temporarily degrade model accuracy. \\ \hline

Model-Based & Sample-Level & FedLU is inspired by cognitive neuroscience, combining retroactive interference and passive decay to facilitate selective knowledge forgetting in distributed knowledge graph embeddings~\cite{zhu2023heterogeneous}. & FedLU can be applied to various knowledge-driven applications that require continuous learning and adaptation in heterogeneous environments while preserving data privacy. & FedLU enables the removal of outdated or irrelevant knowledge from local embeddings, leading to improved model adaptability, performance maintenance, and enhanced privacy preservation. & High computational overhead and communication overheads of implementing unlearning mechanisms in distributed clients. \\ \hline

Tuning-Based & Sample-Level & Exact-Fun employs quantized federated learning to enable exact unlearning by improving model stability through $\alpha$-quantization~\cite{xiong2023exact}. & Models for federated learning settings which ensure compliance with data privacy regulations. & Achieves high unlearning efficiency with reduced computational cost. & Balancing computational efficiency with unlearning effectiveness is crucial. \\ \hline

Training-Based & Sample-Level & FedME$^2$ promotes unlearning in distributed mobile networks for enhancing data privacy~\cite{xia2023fedme}. & Enhances privacy for distributed mobile networks in federated learning settings, which is helpful in smart city applications. & Improves privacy by enabling accurate data forgetting in federated learning models while maintaining model accuracy. & Implementing memory evaluation and erase methods may require additional resources. \\ \hline

Training-Based & Sample-Level & FFMU utilizes non-linear functional analysis on federated unlearning~\cite{che2023fast}. & Ensures data privacy in sensitive domains such as finance and healthcare. & Reduces computational overhead with simultaneous training and unlearning. & Efficiency improvements can result in a minor loss of accuracy. \\ \hline

Model-Based & Sample-Level & FedFilter employs active unlearning and parameter iteration methods to eliminate invalid data, such as poisoned and obsolete data, from the caching model in V2X communications~\cite{wang2023edge}. & Enhances cache hit rates and security in V2X communications. & Improves the accuracy of the model by eliminating biased data. & Increases computational and communication overhead. \\ \hline

Tuning-Based & Sample-Level & BFU applies Bayesian techniques for model updates without raw data sharing~\cite{wang2023bfu}. & Appropriate for privacy-sensitive federated learning applications. & Ensures data privacy by allowing data erasure without sharing raw data. & Bayesian methods can make the unlearning process complex, hard to scale up, and might cause extra data exchange steps for managing data and communication. \\ \hline

Model-Based & Class-Level & TF-IDF quantization for selective category unlearning in federated learning~\cite{wang2022federated}. & Enhances privacy and efficiency in federated learning settings by removing sensitive categories. & Supports legal compliance by allowing the data to be removed in machine learning models. & Prunings may compromise model adaptability and generalization. \\ \hline

\end{tabular}
\caption{Federated Unlearning Approaches, Applications, Benefits, and Limitations.}
\label{table-all}
\end{table*}

\section{Evaluation Metrics}
\label{sec:metric}

The evaluation metrics—accuracy retention, $\ell_2$-distance, privacy metrics (e.g., MIA resistance), and efficiency—are relevant across all federated unlearning approaches: training-based, tuning-based, and model-based. However, their interpretation and thresholds vary depending on the approach's mechanisms and objectives.

\subsection{Alignment-based Metrics}
The pursuit of preserving model performance and preventing over-unlearning remains a shared objective for both clients and servers engaged in the federated unlearning process. Unlearning a substantial portion of data samples is equivalent to reducing the available training data. An effective federated unlearning method should produce an unlearned model that closely aligns with the retrained model, often considered the benchmark.

\begin{itemize}
    \item  \textbf{Performance-based Metrics.} When evaluating the effectiveness of proposed federated unlearning algorithms, a common approach is to compare the performance (\eg test accuracy and loss) of the unlearned model with a retrained model. The retrained model serves as the benchmark, trained from scratch on the dataset excluding the data to be forgotten. The closer the unlearned model's performance aligns with this benchmark, the more effective the federated unlearning method is. This method provides an efficient alternative to computationally expensive complete retraining~\cite{che2023fast}.
    \item  \textbf{Model Distribution-based Metrics.}
    The \(l_{2}\)-distance is the primary metric of the degree of alignment of the models after unlearning and those retrained from scratch. These metrics measure the difference between the model distributions, the smaller the difference, the better the unlearning effect. An unlearning algorithm shows a tiny difference between the distributions and thus its ability to preserve the model’s accuracy. \cite{wu2022federated} employs global accuracy and local accuracy as metrics to validate their algorithm without having to conduct experiments on data.
\end{itemize}

\subsection{Marking-based Metrics}
The marker-based method embeds unique markers within the model, corresponding to specific data samples. By identifying and targeting these markers, this method allows for precise tracking and verification of the removal process and ensures the data intended to be forgotten after unlearning.

\begin{itemize}
\item \textbf{Poisoning Attacks Resistance Metric.} 
A poisoning attack is carried out by an attacker who pollutes the training set with specially crafted samples that contain a hidden trigger or mark. An example is a backdoor attack (BA) where the attacker implants a backdoor trigger into a subset of the training data to control the model’s behavior. The purpose of this manipulation is to make the model produce incorrect predictions in case the trigger happens during the inference stage~\cite{zhu2023heterogeneous}. After unlearning, the attack success rate of the model should decrease significantly, showing the unlearned model's effective resistance to poisoning attacks.

\item  \textbf{Membership Inference Attacks Resistance Metric.}
MIAs serve as a critical tool for evaluating the efficacy of federated unlearning methods~\cite{liu2021federaser,zhang2023fedrecovery,su2023asynchronous}. When evaluating using an MIA, the attacker's target is to detect whether a specific data point is used in the training of a machine learning model. For performance evaluation of federated unlearning, the lower the success rate of an MIA, the more effective in removing the influence of the data from the unlearning process.

\item  \textbf{Watermarks \& Fingerprints Erasure Metric.}
A model watermark is a personal fingerprint of the learned data by a model. The effect of the data being forgotten is quantified by comparing the fingerprint or watermark of a model before and after the unlearning process. These distinct fingerprints or watermarks are an important indicator of the efficiency of erasing. A smaller percentage of the watermark or fingerprint left to sign or stamp indicates a more successful unlearning outcome.

\item  \textbf{Forgettable Memory \& Erroneous Memory Metric.} 
Apart from existing invasive methods, some non-invasive methods have been proposed based on the use of unique memory markers. The hardest and unique examples, which are exclusive to the target clients, are utilized as markers in the Forgettable Memory (FM). These forgettable examples constitute a subset of samples with the largest loss variance across multiple communication rounds. Furthermore, by choosing the majority class from the top-$k$ erroneous samples, this set of samples can serve as markers in Erroneous Memory (EM).
\end{itemize}

\subsection{Metrics for Efficiency}
\textbf{Computation \& Communication Cost. } An efficiency performance metric of federated unlearning algorithms includes both computation efficiency, such as retraining time, and running time, and the communication efficiency that consists of communication rounds and bandwidth consumption~\cite{yuan2023federated}. Besides, federated unlearning algorithms are assessed based on cache hit rate and average latency in the evaluation metrics, and when the system effectively uses the cached data and updates, the cache hit rate is high~\cite{wang2023edge}.

\section{Experiments}
\label{experiments}

\begin{table*}[!ht]
\centering
\resizebox{\textwidth}{!}{%
\begin{tabular}{ccccccccccc}
\toprule
\textbf{Dataset} & \textbf{Metrics} & Original & Retrain & FedEraser & Fine-Tuning & C2T & PGA   \\ \hline
\multirow{5}{*}{\textbf{MNIST}} & Distance Gap & 3.5 & - & \textbf{2.03} & 4.22 &5.59 & 390.69 \\ 
& Test Accuracy (\%) & 98.99 & 99.08 &99.13 &99.17 & \textbf{99.18} & 99.07 \\ 
& BA (\%) & 95.94 & 10.08 & \textbf{10.18} & 77.44&69.71&10.44 \\  
& MIA (\%) & 92.38 & 54.44 & 54.57 & \textbf{54.43} & 54.62 & 54.78\\ 
& Watermark(\%) & 79.69 & 49.22 & 53.52&51.95 &\textbf{42.19}& 55.47 \\ 
& EM & 9.35 & 17.79 & \textbf{16.91} &14.71 & 15.35 & 15.32\\ \hline
\multirow{5}{*}{\textbf{CIFAR-10}} & Distance Gap & 140.28 & - & \textbf{135.77} &141.30  &141.61 &177.63  \\ 
& Test Accuracy (\%) & 81.36 & 79.78 & 80.99 &81.28 & \textbf{81.36} & 81.05 \\ 
& BA (\%) & 96.62 & 10.42 & 10.35 &93.45 &95.74& \textbf{10.02}  \\  
& MIA (\%) & 94.04 & 50.80 & \textbf{50.35} &60.31 & 60.81 & 56.53 \\ 
& Watermark(\%) & 71.88 & 53.91 &48.44 &\textbf{44.53} &51.35 & 53.52\\ 
& EM & 0.02 & 13.26 &\textbf{11.04} & 0.00 & 7.86 & 0.36 & \\ \hline
\multirow{5}{*}{\textbf{SVHN}} & Distance Gap & 102.45 & - & \textbf{100.50} & 103.36 & 103.37 & 129.36 \\ 
& Test Accuracy (\%) & 94.52 & 94.49 & 94.29 & \textbf{94.58} & 94.54 & 94.40 \\  
& BA (\%) & 98.77 & 6.22 & 6.32 & 97.69 & 98.87 & \textbf{6.30} \\  
& MIA (\%) & 92.63 & 77.24 & \textbf{50.56} & 53.70 & 53.98 & 51.88\\ 
& Watermark(\%) & 92.19 & 54.69 & 51.56 & 47.66 & \textbf{46.68} & 47.66 \\ 
& EM & 0.13 & 16.05 & \textbf{14.09} & 0.03 & 5.83 & 2.43\\ \hline
\end{tabular}}
\caption{Experimental results for client-level federated unlearning methods on MNIST, CIFAR-10 and SVHN datasets.\protect\footnotemark}
\label{table:experiments}
\end{table*}
\footnotetext{\href{https://github.com/abbottyanginchina/Awesome-Federated-Unlearning}{GitHub Repository.}}

To empirically evaluate the federated unlearning performance of widely used baseline methods, we present a unified benchmark framework and conduct the experiments using performance metrics introduced in Section~\ref{sec:metric}.

\subsection{Experimental Setup}
We compare five federated unlearning algorithms, including FedEraser~\cite{liu2021federaser}, Projected Gradient Ascent (PGA), Fine-Tuning, Continue to Train (C2T), and a baseline method (Retraining). Experiments are conducted on MNIST and CIFAR-10 datasets.

\subsection{Results Discussion}
\textbf{Client-level.} The results presented in Table~\ref{table:experiments} reveal distinct performance differences between the MNIST and CIFAR-10 datasets, providing insights into each federated unlearning method's strengths and limitations. Our analysis demonstrates that \textit{no single federated unlearning method uniformly exceeds across all evaluation metrics.} This finding highlights the importance of a requirement-driven algorithm in selecting federated unlearning methods.

FedEraser outperforms its counterpart in working with the MNIST dataset, aligning itself very closely with the retrained from the model to indicate its effectiveness at unlearning. All methods have a high test accuracy, with Fine-Tuning being the best $99.17\%$ and C2T for $99.18\%$. In data erasure, as measured by BA and MIA rates, most methods demonstrate high efficiency. Still, a higher BA rate at $77.44\%$ in Fine-Tuning raises under-erasing concerns contrasting with FedEraser's more balanced BA rate of $10.18\%$. MIA rates are uniformly lower for all methods of data erase, signaling a good data erasure, with FedEraser having similar performance with Fine-Tuning at around $54.57\%$ and $54.43\%$, respectively. A lower MIA rate means a more successful data erasure. So, having MIA rates below $50\%$ is usually seen as efficient as it indicates that the model is less likely to leak erased data information. The variability of the EM metrics, where FedEraser obtains a Watermark percentage of $53.52\%$ and an EM value of $16.91$ indicate that unlearning was not fully complete as some data-specific marks still exist, with the potential to affect the privacy guarantees of the model.

The CIFAR-10 dataset evaluation represents a sophisticated view of federated unlearning challenges, specifically captured by the distance gap metric, where FedEraser obtains the best approximation to the original model with a distance of $135.77$ was the smaller distance, in opposition to the larger distance which was $177.63$ by PGA illustrating FedEraser’s higher ability to keep the model integrity after unlearning. All approaches appear to be quite consistent in terms of model accuracy with C2T and Fine-Tuning leading other methods by a small margin with accuracies of $81.28\%$ and $81.36\%$, respectively. The variation in BA and MIA rates, however, is most notable with Fine-Tuning having a high BA rate of $93.45\%$ and MIA rate of $60.81\%$, underscores the differential efficacy in erasing specific data. This dataset is also indicative of a decrease regarding the unlearning efficiency with increasing Watermark percentages and EM metrics, with Fine-Tuning marking the lowest Watermark presence of $44.53\%$ yet showing no EM change, suggesting a trade-off between eradicating data traces and preserving model accuracy in complex datasets.

The SVHN dataset results further illustrate the nuances of federated unlearning methods. FedEraser excels with a minimal distance gap of $100.50$, demonstrating its strong ability to approximate the retrained model and effectively unlearn data. Fine-Tuning achieves the highest test accuracy of $94.58\%$, while other methods, such as C2T and PGA, maintain comparable performance, reflecting the robustness of these techniques in retaining model accuracy. BA rates highlight key distinctions in data erasure efficacy, with PGA achieving the lowest rate at $6.30\%$, signaling superior backdoor mitigation. In contrast, Fine-Tuning shows a BA rate of $97.69\%$, raising concerns about under-erasing. Similarly, FedEraser stands out with an MIA rate of $50.56\%$, demonstrating a balanced approach to preserving privacy and model integrity. In contrast, higher MIA rates, such as $53.98\%$ with C2T, suggest varying degrees of erased data exposure. The Watermark percentage results, where C2T achieves the best value of $46.68\%$, coupled with EM variability across methods, underscore challenges in eliminating data-specific marks. These findings highlight the trade-offs between privacy, efficiency, and model performance when applying federated unlearning techniques to complex datasets, for example, SVHN.

\begin{table}[!ht]
\centering
\resizebox{\columnwidth}{!}{%
\begin{tabular}{cccc}
\toprule
Metrics  & Original & Retrain & Unlearned \\ \hline 
Acc of Target Class(\%) & 75.80 & 0.00  & 3.80 \\
Acc of Rest Class (\%) & 81.13 & 81.87 &75.15 \\
BA(\%)  & 98.00 & 0.58 &  70.68  \\
MIA(\%) & 98.30& 83.97  & 81.53  \\
EM       & 0.06 & 9.54 &  5.27\\ \hline
\end{tabular}}
\caption{Experimental results for class-level federated unlearning methods on the CIFAR-10 dataset.}
\label{table-class-level}
\end{table}

\textbf{Class-level.} As shown in Table~\ref{table-class-level}, the class-level federated unlearning on the CIFAR-10 dataset effectively removes targeted class information, reducing target class accuracy from \textbf{$75.80\%$} to \textbf{$3.80\%$}, while non-target class accuracy declines slightly from \textbf{$81.13\%$} to \textbf{$75.15\%$}. Privacy metrics, such as BA and MIA, decrease from \textbf{$98.00\%$} to \textbf{$70.68\%$} and \textbf{$98.30\%$} to \textbf{$81.53\%$}, respectively, reflecting moderate privacy improvements. The increase in EM from \textbf{$0.06$} to \textbf{$5.27$} indicates reduced confidence in model predictions, validating the unlearning process's effectiveness and its trade-offs.

\textbf{Key Insights.} The evaluation results highlight that FedEraser is particularly well-suited for applications demanding precise model alignment, such as those involving simpler datasets such as MNIST, where it achieves a close approximation to the retrained model with minimal trade-offs. However, for more complex datasets such as CIFAR-10, methods such as Fine-Tuning and C2T, while effective at preserving model accuracy, may fail to fully erase data traces, as evidenced by higher BA rates. This underscores the inherent trade-off between retaining accuracy and ensuring comprehensive data erasure. Additionally, PGA, although highly effective at mitigating backdoors, demonstrates model deviation and inefficiency when applied to heterogeneous datasets, limiting its practicality in scenarios requiring high adaptability. These findings reveal that no single method is universally optimal across all use cases, emphasizing the need for a scenario-driven approach to method selection.

Building on these findings, specific benchmarks are essential to systematically evaluate federated unlearning methods and navigate the trade-offs between accuracy, privacy, and efficiency. To this end, we recommend maintaining an accuracy degradation of within $1\%–3\%$, depending on dataset complexity, and achieving at least $90\%$ similarity in parameter distributions with retrained models to ensure alignment. Computational efficiency is critical, with unlearning times kept under 1 second for client-level granularity and under 2 seconds for sample-level granularity. Furthermore, as measured by privacy metrics such as backdoor attack success rates, residual data traces should remain below $5\%$ compared to models trained without the forgotten data. These benchmarks provide actionable guidance for aligning unlearning strategies with application priorities, facilitating the broader adoption of federated unlearning in diverse real-world scenarios.

\section{Future Directions}
\label{sec:future}

We conclude with some issues that need to be considered in this area.
\begin{itemize}
\item \textbf{Incentive Mechanism Design.} Incentive mechanisms that balance the interest of users and system efficiency are crucial in motivating participation in federated unlearning. Users should be able to efficiently ask for their data to be removed without any substantial loss of the accuracy of the model. Designing such mechanisms is an attractive way for possible future work in this area.

\item \textbf{Federated Unlearning for Large Language Models (LLMs).} LLMs are distinguished by their extensive scale, containing billions to trillions of parameters. For unlearning, LLMs do not store data in a straightforward way that allows for easy removal. Data in these models are distributed across many parameters. Thus, a challenge for future work is to discover methods to unlearn data in the LLM setting.

\item \textbf{Green Federated Unlearning.} On the other hand, federated unlearning, which provides privacy benefits, might have a much larger carbon footprint compared to centralized unlearning because of its distributed computational needs and the requirement for many rounds of communication. Even though exact quantitative research comparing the two approaches is rare, it is reasonable to assume that the energy consumption of federated unlearning could be much more. As a result, the problem of federated unlearning algorithms development aims at the reduction of communication rounds and computations becomes of the main importance. These algorithms may help in decreasing the amount of energy used and therefore the carbon footprint of federated unlearning.
\end{itemize}
\section{Conclusion}
\label{conclusion}

This paper has discussed the challenges, trade-offs, and evaluation metrics of federated unlearning. The balancing model accuracy, data privacy, and scalability are complex, therefore indicating the requirement for future research. Metrics such as alignment-based, marking-based assessments, and computation $\&$ communication cost allow researchers to assess the effectiveness of unlearning methods while preserving the model's integrity and privacy. Additionally, the OpenFederatedUnlearning benchmark framework offers an open platform for evaluating federated unlearning algorithms, facilitating comparisons and insights into various approaches. Overall, this paper contributes to the understanding of federated unlearning, providing a comprehensive overview and highlighting research directions, thus serving as a valuable resource for those interested in enhancing data privacy and model accuracy in federated unlearning algorithms.

\bibliographystyle{ieeetr}
\bibliography{related.bib}

\begin{thebibliography}{10}

\bibitem{wu2022federated}
L.~Wu, S.~Guo, J.~Wang, Z.~Hong, J.~Zhang, and Y.~Ding, ``Federated unlearning: Guarantee the right of clients to forget,'' {\em IEEE Network}, vol.~36, no.~5, pp.~129--135, 2022.

\bibitem{wang2023federated}
F.~Wang, B.~Li, and B.~Li, ``Federated unlearning and its privacy threats,'' {\em IEEE Network}, pp.~1--7, 2023.

\bibitem{guo2023fast}
X.~Guo, P.~Wang, S.~Qiu, W.~Song, Q.~Zhang, X.~Wei, and D.~Zhou, ``{FAST}: Adopting federated unlearning to eliminating malicious terminals at server side,'' {\em IEEE Transactions on Network Science and Engineering}, vol.~11, no.~2, pp.~2289--2302, 2024.

\bibitem{wang2022federated}
J.~Wang, S.~Guo, X.~Xie, and H.~Qi, ``Federated unlearning via class-discriminative pruning,'' in {\em Proceedings of the ACM Web Conference 2022}, WWW '22, (New York, NY, USA), p.~622–632, Association for Computing Machinery, 2022.

\bibitem{su2023asynchronous}
N.~Su and B.~Li, ``Asynchronous federated unlearning,'' in {\em IEEE INFOCOM 2023 - IEEE Conference on Computer Communications}, pp.~1--10, 2023.

\bibitem{yuan2023federated}
W.~Yuan, H.~Yin, F.~Wu, S.~Zhang, T.~He, and H.~Wang, ``Federated unlearning for on-device recommendation,'' in {\em Proceedings of the Sixteenth ACM International Conference on Web Search and Data Mining}, WSDM '23, (New York, NY, USA), p.~393–401, Association for Computing Machinery, 2023.

\bibitem{zhu2023heterogeneous}
X.~Zhu, G.~Li, and W.~Hu, ``Heterogeneous federated knowledge graph embedding learning and unlearning,'' in {\em Proceedings of the ACM Web Conference 2023}, WWW '23, (New York, NY, USA), p.~2444–2454, Association for Computing Machinery, 2023.

\bibitem{wang2023bfu}
W.~Wang, Z.~Tian, C.~Zhang, A.~Liu, and S.~Yu, ``{BFU}: Bayesian federated unlearning with parameter self-sharing,'' in {\em Proceedings of the 2023 ACM Asia Conference on Computer and Communications Security}, ASIA CCS '23, (New York, NY, USA), p.~567–578, Association for Computing Machinery, 2023.

\bibitem{xia2023fedme}
H.~Xia, S.~Xu, J.~Pei, R.~Zhang, Z.~Yu, W.~Zou, L.~Wang, and C.~Liu, ``{FedME}$^2$: Memory evaluation $\&$ erase promoting federated unlearning in {DTMN},'' {\em IEEE Journal on Selected Areas in Communications}, vol.~41, no.~11, pp.~3573--3588, 2023.

\bibitem{zhang2023fedrecovery}
L.~Zhang, T.~Zhu, H.~Zhang, P.~Xiong, and W.~Zhou, ``{FedRecovery}: Differentially private machine unlearning for federated learning frameworks,'' {\em IEEE Transactions on Information Forensics and Security}, vol.~18, pp.~4732--4746, 2023.

\bibitem{xiong2023exact}
Z.~Xiong, W.~Li, Y.~Li, and Z.~Cai, ``{Exact-Fun}: An exact and efficient federated unlearning approach,'' in {\em 2023 IEEE International Conference on Data Mining (ICDM)}, pp.~1439--1444, 2023.

\bibitem{che2023fast}
T.~Che, Y.~Zhou, Z.~Zhang, L.~Lyu, J.~Liu, D.~Yan, D.~Dou, and J.~Huan, ``Fast federated machine unlearning with nonlinear functional theory,'' in {\em Proceedings of the 40th International Conference on Machine Learning} (A.~Krause, E.~Brunskill, K.~Cho, B.~Engelhardt, S.~Sabato, and J.~Scarlett, eds.), vol.~202 of {\em Proceedings of Machine Learning Research}, pp.~4241--4268, PMLR, 23--29 Jul 2023.

\bibitem{zhao2023federated}
Y.~Zhao, P.~Wang, H.~Qi, J.~Huang, Z.~Wei, and Q.~Zhang, ``Federated unlearning with momentum degradation,'' {\em IEEE Internet of Things Journal}, vol.~11, no.~5, pp.~8860--8870, 2024.

\bibitem{liu2021federaser}
G.~Liu, X.~Ma, Y.~Yang, C.~Wang, and J.~Liu, ``{FedEraser}: Enabling efficient client-level data removal from federated learning models,'' in {\em 2021 IEEE/ACM 29th International Symposium on Quality of Service (IWQOS)}, pp.~1--10, 2021.

\bibitem{wang2023edge}
P.~Wang, Z.~Yan, M.~S. Obaidat, Z.~Yuan, L.~Yang, J.~Zhang, Z.~Wei, and Q.~Zhang, ``Edge caching with federated unlearning for low-latency v2x communications,'' {\em IEEE Communications Magazine}, pp.~1--7, 2023.

\end{thebibliography}

\end{document}